\documentclass[letterpaper]{article} % DO NOT CHANGE THIS
\usepackage{aaai25}  % DO NOT CHANGE THIS
\usepackage{times}  % DO NOT CHANGE THIS
\usepackage{helvet}  % DO NOT CHANGE THIS
\usepackage{courier}  % DO NOT CHANGE THIS
\usepackage[hyphens]{url}  % DO NOT CHANGE THIS
\usepackage{graphicx} % DO NOT CHANGE THIS
\usepackage{natbib}  % DO NOT CHANGE THIS AND DO NOT ADD ANY OPTIONS TO IT
\usepackage{caption} % DO NOT CHANGE THIS AND DO NOT ADD ANY OPTIONS TO IT
\usepackage{algorithm}

\usepackage{hyperref}

\usepackage{algorithmicx}
\usepackage{algpseudocode}
\usepackage{amsmath}
\usepackage{amssymb}
\usepackage{booktabs}
\usepackage{multirow}
\usepackage{bm}
\usepackage[table]{xcolor}
\usepackage{pifont}

\usepackage[misc]{ifsym}

\def \adapterlogo {\raisebox{-0.1\height}{\includegraphics[height=0.95\baselineskip]{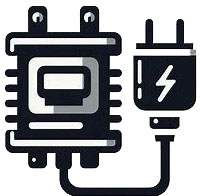}}}

\usepackage{newfloat}
\usepackage{listings}
\DeclareCaptionStyle{ruled}{labelfont=normalfont,labelsep=colon,strut=off} % DO NOT CHANGE THIS
\floatstyle{ruled}
\newfloat{listing}{tb}{lst}{}
\floatname{listing}{Listing}
\nocopyright

\title{\adapterlogo{} 5\%$>$100\%: Breaking Performance Shackles of Full Fine-Tuning \\on Visual Recognition Tasks}

\author{
    Dongshuo Yin\textsuperscript{\rm 1}\equalcontrib, 
    Leiyi Hu\textsuperscript{\rm 2}\equalcontrib, 
    Bin Li\textsuperscript{\rm 3}, 
    Youqun Zhang\textsuperscript{\rm 3}, 
    Xue Yang\textsuperscript{\rm 4}\thanks{Corresponding author.}
}
\affiliations{
    \textsuperscript{\rm 1}BNRist, Department of Computer Science and Technology, Tsinghua University\\
    \textsuperscript{\rm 2}University of Chinese Academy of Sciences
    \textsuperscript{\rm 3}Alibaba Group
    \textsuperscript{\rm 4}Shanghai Jiao Tong University
    dongshuoyin@gmail.com, 
    hu\_leiyi@163.com,\\
    \{zhuyi.lb, youqun.zhangyq\}@alibaba-inc.com, yangxue-2019-sjtu@sjtu.edu.cn

}

\usepackage{bibentry}
\begin{document}

\maketitle

\begin{abstract}
Pre-training \& fine-tuning can enhance the transferring efficiency and performance in visual tasks. Recent delta-tuning methods provide more options for visual classification tasks. Despite their success, existing visual delta-tuning art fails to exceed the upper limit of full fine-tuning on challenging tasks like object detection and segmentation. To find a competitive alternative to full fine-tuning, we propose the \textbf{M}ulti-c\textbf{o}g\textbf{n}itive Visual \textbf{A}dapter \textbf{(Mona)} tuning, a novel adapter-based tuning method. First, we introduce multiple vision-friendly filters into the adapter to enhance its ability to process visual signals, while previous methods mainly rely on language-friendly linear filters. Second, we add the scaled normalization layer in the adapter to regulate the distribution of input features for visual filters. To fully demonstrate the practicality and generality of Mona, we conduct experiments on multiple representative visual tasks, including instance segmentation on COCO, semantic segmentation on ADE20K, object detection on Pascal VOC, oriented object detection on DOTA/STAR, and image classification on three common datasets. Exciting results illustrate that Mona surpasses full fine-tuning on all these tasks, and is the only delta-tuning method outperforming full fine-tuning on the above various tasks. For example, Mona achieves 1\% performance gain on the COCO dataset compared to full fine-tuning. Comprehensive results suggest that Mona-tuning is more suitable for retaining and utilizing the capabilities of pre-trained models than full fine-tuning. The code will be released at
\href{https://github.com/Leiyi-Hu/mona}{https://github.com/Leiyi-Hu/mona}
.
\end{abstract}

\section{Introduction}
\label{sec:intro}

Pre-training \& fine-tuning paradigm \cite{wang2022pre} can perform impressive transfer learning between homo-modal tasks, as has been demonstrated in computer vision (CV) \cite{liu2021swin, fang2023eva} and natural language processing (NLP) \cite{tufano2022using, min2023recent, tinn2023fine}. Pre-trained models are often trained by well-resourced and experienced teams with large amounts of clean data \cite{yin2023parameter}. Exceptional pre-trained models can help hardware- and data-limited teams save plenty of training costs and train well-performing deep models on new tasks \cite{sarasaen2021fine, amisse2021fine, too2019comparative, kading2016fine}. In the era of large models, the efficiency of tuning pre-trained models is an important issue. Full fine-tuning has been widely used with great success in CV tasks, which tunes all parameters in the pre-trained backbone as well as additional task-specific heads/necks during the training process. Many impressive CV art push the limit of visual tasks through pretraining \& full fine-tunning. However, is full fine-tuning still the best way to fine-tune visual tasks now?

\begin{figure}[tb]
  \centering
  \includegraphics[scale=.76]{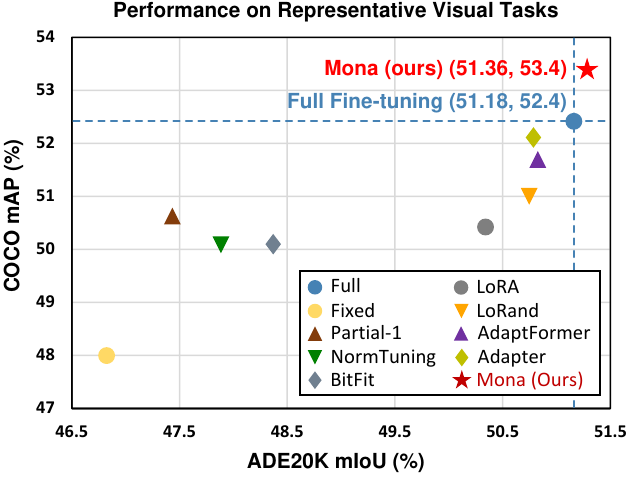}
  \caption{\textbf{Comparisons of our method with full fine-tuning and recent delta-tuning art on representative visual tasks.} Blue dashed line is the performance of full fine-tuning on ADE20K and COCO. The proposed Mona outperforms full fine-tuning on representative visual tasks, which promotes the upper limit of previous delta-tuning art. The results demonstrate that the adapter-tuning paradigm can replace full fine-tuning and achieve better performance in common visual tasks. Full fine-tuning may no longer be the only preferred solution for transfer learning in the future.} \label{fig:first}
\end{figure}

Apart from full fine-tuning, Delta tuning \cite{ding2023parameter, hu2022sparse} has recently attracted attention in NLP and CV tasks. Delta tuning comes from NLP, which tunes only part of the backbone network or extra lightweight structures for efficient transfer learning \cite{ding2023parameter}. Delta tuning methods generally fix most backbone parameters and achieve comparable or even better performance than full fine-tuning on simple tasks (including classification tasks in NLP \cite{zhou2022making, rathnayake2022adapter} and CV \cite{jia2022visual, liu2022polyhistor, he2022parameter, chen2022adaptformer}). VPT \cite{jia2022visual} is the first to explore the potential of prompt-tuning on visual classification tasks. LoRand \cite{yin20231} pioneers adapter-tuning on dense predictions and reduces the gap between delta tuning and full fine-tuning on visual tasks. However, existing methods cannot outperform full fine-tuning on visual recognition tasks, including semantic and instance segmentation.

To challenge the dominance of full fine-tuning in CV, we propose Mona-tuning, a novel tuning paradigm based on \textbf{M}ulti-c\textbf{o}g\textbf{n}itive visual \textbf{a}dapters \textbf{(Mona)}. We analyse recent art and summarise two issues in existing visual adapters. First, the designs of existing CV adapters \cite{liu2022polyhistor, he2022parameter, chen2022adaptformer} follow linear adapters in NLP \cite{houlsby2019parameter}. In fact, visual tasks process visual signals, which are significantly different from linguistic signals and have unique 2D convolutional operations \cite{gu2018recent, li2021survey, albawi2017understanding}. Our experiments show that convolution-based filters can better transfer visual knowledge from pre-trained models to other tasks, so we propose a practical convolution-based adapter for visual tasks. Second, most existing adapters compress the upstream features with a single linear layer \cite{liu2022polyhistor, he2022parameter}. Previous works claim that models have different cognition of features at different filter scales \cite{ozturk2018convolution, chansong2021impacts, agrawal2020using}. Thus, we employ multiple convolutional filters behind the adapter's reduction layer to enhance the cognitive abilities of the adapters. We demonstrate the generality and superiority of Mona-tuning on plenty of representative visual tasks, including image classification, object detection, semantic segmentation, instance segmentation, and oriented object detection \cite{yang2019scrdet,yang2021r3det}. We employ the SwinTransformer \cite{liu2021swin} series trained on ImageNet-22k \cite{deng2009imagenet} as pre-trained models. Extensive experiments indicate that the proposed method outperforms the traditional full fine-tuning paradigm both on simple image classification tasks and complex visual tasks. For example, Mona-tuning outperforms full fine-tuning on the COCO dataset \cite{lin2014microsoft} by 1\% mAP. The results suggest that full fine-tuning may no longer be the optimal choice for visual tasks. 
% Adapter-based tuning is a better-performing and parameter-efficient paradigm for visual transfer learning. Moreover, 
As far as we known, Mona is the only Adapter-based tuning method that surpasses full fine-tuning on semantic segmentation, instance segmentation, and oriented object detection. Figure \ref{fig:first} illustrates the superiority of the proposed method on the challenging instance segmentation and semantic segmentation tasks. Our contributions can be three-fold:

\begin{itemize}
\item[$\bullet$] We demonstrate that the adapter-based tuning can surpass full fine-tuning on visual tasks, and perform better than full fine-tunning with fewer new parameters.
\item[$\bullet$] We propose Mona-tuning, a novel and practical training paradigm based on multi-cognitive visual adapters (Mona). Mona employs vision-friendly filters to optimise traditional linear adapters and improve the transferring efficiency of visual pre-trained knowledge through multiple cognitive perspectives.
\item[$\bullet$] Extensive experiments demonstrate that Mona-tuning outperforms full fine-tuning and other recent art on representative visual tasks, including image classification, object detection, semantic segmentation, instance segmentation, and oriented object detection.
\end{itemize}

\section{Related Work}
\subsection{Delta-tuning}
The development of large models has produced dramatic shocks throughout artificial intelligence \cite{floridi2020gpt, touvron2023llama, kirillov2023segment}. The efficiency of transfer learning attracts researchers' interest \cite{chen2023sam, liu2022polyhistor, he2022parameter, chen2022adaptformer}. Delta tuning \cite{houlsby2019parameter, hu2021lora, hu2022sparse, ding2023parameter, si2024flora} (or parameter efficient fine-tuning, PEFT) is dedicated to improving the efficiency of fine-tuning. Delta-tuning methods can be divided into three groups \cite{ding2023parameter}. The first group fixes most of the parameters in the pre-trained backbone and fine-tune a small number of them, e.g., BitFit \cite{zaken2021bitfit} tunes bias, Norm Tuning \cite{giannou2023expressive} tunes norm layers, and Partial-1 \cite{yosinski2014transferable} only tunes the last block. The second group reparameterises some parameters in the pre-trained model, e.g. the LoRA \cite{hu2021lora} optimises low-rank subspaces. The third group fixes the pre-trained backbone's original parameters and adds additional trainable structures, including prompt series \cite{jia2022visual, liu2022p, zhu2023prompt} and adapter series \cite{sung2022vl, chen2022adaptformer, he2023parameter}. Our experiments compare Mona with these three groups.

\subsection{Computer Vision Meets Delta-tuning}
Although derived from NLP, delta tuning is also explored in CV. VPT \cite{jia2022visual} is the first to introduce delta-tuning (prompt-tuning) to visual classification tasks. \cite{chen2022vision} adds adapters to a trainable backbone to improve performance rather than parameter efficiency. AdaptFormer \cite{chen2022adaptformer} designs a parallel adapter structure to improve delta-tuning performance on visual classification. KAdaptation \cite{he2023parameter} optimises the adapter through the Kronecker product. The above art is the pioneer in visual tasks, revealing the potential of delta-tuning on visual classification. LoRand \cite{yin20231} brings impressive performance on dense prediction tasks via multi-branch low-rank adapters but still cannot surpass full fine-tuning on all visual recognition tasks. Recent art indicates that delta-tuning cannot completely replace full fine-tuning on vision tasks. Therefore, we propose Mona-tuning, an alternative to full fine-tuning for more visual tasks, which outperforms full fine-tuning in both new parameter sizes and performance.

\section{Methods}
In this section, we present the proposed method in four parts, including the adapter-tuning, Mona, and parameter analysis.

\subsection{Adapter-tuning} \label{adapter}
Previous work \cite{yin20231} discussed adapter fine-tuning, and we briefly introduce related concepts here. Full fine-tuning updates all parameters in the pre-trained backbone, while adapter-tuning fixes the pre-trained parameters and updates the parameters in adapters. For dataset $D=\left\{(x_i,y_i)\right\}^N_{i=1}$, the optimization process of full fine-tuning and adapter-tuning can be expressed as Eq. \ref{eq1} and Eq. \ref{eq2}:

\begin{equation}
\label{eq1}
	\theta\gets\underset{\theta}{{\arg\min} \, loss(D,\theta)},
\end{equation}
\begin{equation}
\label{eq2}
\omega\gets\underset{\omega}{{\arg\min} \, loss(D,\theta_F, \omega)},
\end{equation}
where $loss$ is the training loss, $\theta$ represents parameters of the whole framework, and $\theta_F$ is the fixed parameters in adapter-tuning. $\omega$ represents updated parameters in adapter-tuning, including parameters in adapters and outside the backbone.

\begin{figure}[tb]
	\centering
	\includegraphics[scale=.65]{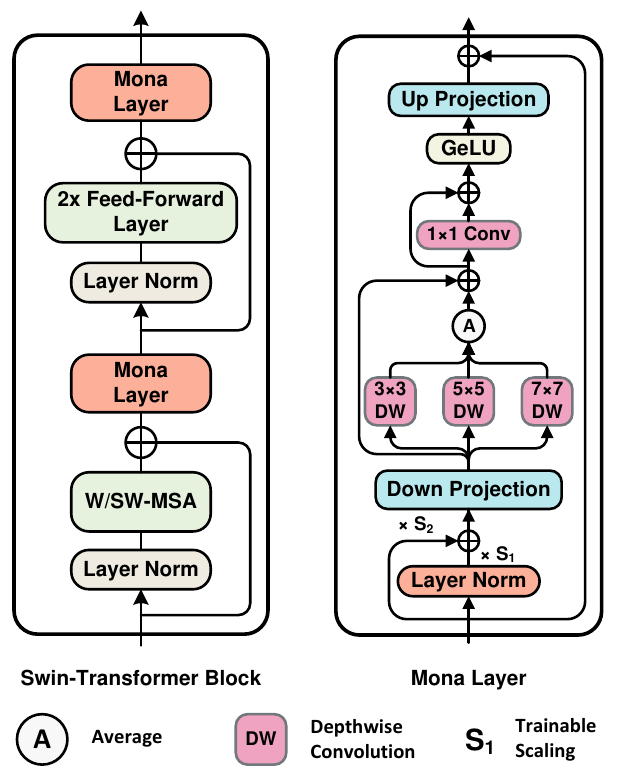}
	\caption{\textbf{Left:} The proposed Mona-tuning. We add Mona after MSA and MLP in each SwinBlock. The proposed method fixes the parameters of pre-trained layers and updates the parameters of Mona. \textbf{Right:} Details of Mona. Mona has a scaled LayerNorm before the down projection. A multi-cognitive convolutional filter group and an aggregation filter are behind the down projection. We add skip-connections at four places inside Mona to strengthen its adaptation capabilities. Mona enables the adapter-based fine-tuning paradigm to outperform full fine-tuning in typical visual tasks comprehensively.} \label{fig:adapter}
% \vspace{-10pt}

\end{figure}

\subsection{Mona}
\label{mona}

Typical linear adapters suffer from two problems when applied to visual tasks. First, fixed layer parameters cannot be fine-tuned to match the data distribution of new tasks, resulting in a biased feature distribution passed to the adapter. Therefore, it is important for the adapter to optimize its input distribution from fixed layers. Second, Vanilla adapter \cite{houlsby2019parameter} is designed for natural language signals and is not optimized for visual signals. Previous CV adapter art \cite{jia2022visual, liu2022polyhistor, he2022parameter, chen2022adaptformer} is based on linear filters (mainly including down projection, nonlinear activation, up projection, and skip connections), which is not efficient for transferring visual knowledge. To address these two issues, we perform input optimization and design multi-cognitive visual filters.

\noindent \textbf{Input Optimization.} 
% The input features of adapter come from fixed layers, which cannot adjust their parameter spaces according to new tasks' sample space. Therefore, how to efficiently adjust the model's parameter space is a key to designing an adapter. Vanilla adapter directly downscales inputs from fixed layers, which is not a good choice for visual tasks. Thus, 
We enable Mona to adjust the input distributions and the proportion of inputs from the fixed layers. Specifically, we add a norm layer and two learnable weights, $s_1$ and $s_2$, to the top end of Mona to adjust the input distribution. Previous work indicates that normalization \cite{xu2019understanding} helps to stabilize the forward input distribution and the backpropagated gradient. We find in practice that LayerNorm (LN) \cite{xu2019understanding} is better than BatchNorm \cite{ioffe2015batch}, so we employ LN in Mona. Figure \ref{fig:adapter} illustrates our design, which can be formulated as:
\begin{equation}
    x_{norm}=s_1 \cdot |x_0|_{LN}+s_2 \cdot x_0,
\end{equation}
where $|\cdot|_{LN}$ denotes LayerNorm and $x_0$ denotes the original input of Mona.

\noindent \textbf{Multi-Cognitive Visual Filters.} 
% Previous CV adapter art \cite{jia2022visual, liu2022polyhistor, he2022parameter, chen2022adaptformer} is based on linear structures, mainly including down projection, nonlinear activation, up projection, and skip connections. Vanilla adapter \cite{houlsby2019parameter} is for natural language signals and is not optimized for visual signals. Given the limitations of vanilla adapters, we focus on how to enhance the adapter's capability to transfer visual knowledge. 
For visual cognition, human eyes process visual signals from different scales and integrate them for better understanding \cite{koretz1988human, martinez2004role, szegedy2015going}. Adapters should also process upstream features from multiple cognitive perspectives for better performance on downstream tasks. We introduce multiple convolutional filters to Mona to increase the cognitive dimension. Instead of standard convolutions, Depth-Wise Convolutions \cite{szegedy2015going} (DWConv) are employed in Mona to minimize additional parameter sizes. Specifically, the upstream features go through three DWConv filters after down projection. Convolution kernels are 3$\times$3, 5$\times$5 and 7$\times$7. We compute the average results from three filters and aggregate features with a 1$\times$1 convolution. Skip-connections are added to two types of convolutions. We use three depth-wise convolutions with weight $\omega_{dw}^i\in\mathbb{R}^{C_{in}^D\times K_i\times K_i\times C_{out}^D}$ ($i\in{1,2,3}$) for the first multi-filter convolution and a point-wise convolution with weight $\omega_{pw}^i\in\mathbb{R}^{C_{in}^P\times 1\times 1\times C_{out}^P}$ for the second convolution. The above two convolution steps can be formulated as follows:
\begin{equation}
    \begin{aligned}
& f_{dw}=x+avg(\textstyle\sum\nolimits_{i=1}^3 \omega_{dw}^i \hat{\otimes} x),\\
& f_{pw}=x+\omega_{pw} \overline{\otimes} x,
    \end{aligned}
\end{equation}
where $\hat{\otimes}$ and $\overline{\otimes}$ denote depth-wise and point-wise convolution. Then, features are nonlinearized by GeLU and recovered by up projection. The overall calculation process of Mona can be formulated as follows:
\begin{equation}
    x=x_0+U^l\sigma(f_{pw}(f_{dw}(D^l(x_{norm}))),
\end{equation}
where $D^l$ and $U^l$ denote down and up projections of the $l^{th}$ adapter, and $\sigma$ denotes GeLU activation.

\subsection{Parameter Analysis}
\label{param}
The parameters of Mona come from LN, scaling factors, linear layers, DWconv and 1$\times$1 conv. Assuming that the input dimension of the adapter is $m$ and the dimension after down projection is $n$, the parameters of the LN and scaling factors are $2m+2$, the parameters of the two linear layers are $2mn+m+n$, the parameters of the DWConv layer are $(3^2+5^2+7^2)n=83n$, and the PWConv is $n^2$. The total parameter of each Mona module are: 
\begin{equation}
    (2n+3) m+n^2+84n+2.
\end{equation}
For each block, all Mona parameters are: $2\times\left((2n+3)m+n^2+84n+2\right)$. We set the value of $n$ to a constant (64) to reduce parameters in Mona.

% \begin{figure*}[tb]
% 	\centering
% 	\includegraphics[scale=.7]{fig/4-3.pdf}
% 	\caption{\textbf{Design Iterations.} Mona surpasses full fine-tuning on typical visual tasks after many iterations. The first version introduces multi-cognitive visual filters and outperforms previous adapter-based tuning art on visual tasks. We add LNs at different places and incorporate averaging operations to optimise Mona further. In fact, Mona finally succeeds in achieving our desired goal through more than 15 versions.} \label{fig:design}

% % \vspace{-10pt}
% \end{figure*}

\section{Experiments}
We implement sufficient experiments on multiple representative visual tasks to demonstrate the superiority of Mona-tuning. This section includes experimental settings, results, convergence analysis and some ablation experiments. Hyperparameters and detailed settings for training are presented in the Supplementary Material.

\subsection{Datasets}
\label{dataset}
\noindent \textbf{Object Detection.} Pascal VOC 0712 \cite{everingham2015pascal} has 16k/5k training/validation images and is used for object detection tasks. We employ Swin-Large + RetinaNet for training. The evaluation metric for object detection task is the most commonly used AP$_{box}$.

\noindent \textbf{Semantic Segmentation.} ADE20K \cite{zhou2017scene} is the most widely used semantic segmentation dataset containing 20K training and 2K validation images. We employ Swin-Large + UperNet for experiments on semantic segmentation. The evaluation metric is the most commonly used mIoU.

%\noindent \textbf{Semantic Segmentation.} CityScapes \cite{cordts2016cityscapes} is a large-scale semantic segmentation dataset recording street scenes from 50 cities. We employ Swin-Large+UperNet for experiments on semantic segmentation.

\noindent \textbf{Instance Segmentation.} MS COCO \cite{lin2014microsoft} is a representative instance segmentation dataset with 118k training images and 5k validation images. We employ Swin-Base + Cascade Mask RCNN for training. Evaluation metrics for instance segmentation task are AP$_{box}$ and AP$_{Mask}$.

\noindent \textbf{Oriented Object Detection.} Oriented object detection considers angle information in the annotation and inference process, which can effectively improve the performance and efficiency of object detection in fields like remote sensing. This task requires more accurate annotation information and more complex detection models, which is more challenging than horizontal object detection. Two representative remote sensing datasets, DOTA \cite{dota} and STAR \cite{li2024star}, are selected for our experiments. We also experiment with multiple detection frameworks on the more challenging STAR dataset. The meteic here is AP$_{box}$.

\noindent \textbf{Image Classification.} Classification tasks have been well studied in previous art. We also conduct experiments on Oxford 102 Flower \cite{nilsback2008automated}, Oxford-IIIT Pet  \cite{parkhi2012cats}, and VOC 2007 Classification dataset \cite{pascal-voc-2007} to increase the broadness of our experiments. The top-1, top-5, and average accuracy of each method are reported.

\subsection{Pre-trained Models and Toolkits}
\label{pretrained}

The Swin Transformer series \cite{liu2021swin} is employed as the backbone for all experiments. The pre-trained models are trained on ImageNet-22k \cite{deng2009imagenet}, and toolkits like MMDetection \cite{chen2019mmdetection}, MMSegmentation \cite{mmseg2020}, MMRotate \cite{zhou2022mmrotate} and MMClassification \cite{2020mmclassification} are used for verification. The image resolution of the pre-trained task is 224$\times$224. Most tasks employ Swin-Large as the backbone. Backbones for COCO, DOTA, and STAR are Swin-Base, considering the memory consumption of these tasks.
\subsection{Baselines}
\label{basline}

We compare Mona with multiple recent methods. Baselines can be grouped into methods without or with extra structure:
\begin{itemize}
	\item[$\bullet$] Without extra structure:
%\end{itemize}

%\noindent - {\scshape Full}: Update all parameters in the framework.
%
%\noindent - {\scshape Fixed}: Fix the backbone and update other parameters in the framework.
%
%\noindent - {\scshape BitFit \cite{zaken2021bitfit}}: Update bias in backbone and parameters outside of backbone.
%
%\noindent - {\scshape NormTuning \cite{giannou2023expressive}}: Update norm layers in backbone and parameters outside of backbone.
%
%\noindent - {\scshape Partial-1 \cite{yosinski2014transferable}}: Update the last block in the backbone and parameters outside the backbone.

\item[-] {\scshape Full}: Update all parameters in the framework.

\item[-] {\scshape Fixed}: Fix the backbone and update other parameters.

\item[-] {\scshape BitFit} \cite{zaken2021bitfit}: Update bias in backbone and parameters outside of backbone.

\item[-] {\scshape NormTuning} \cite{giannou2023expressive}: Update norm layers in backbone and parameters outside the backbone.

\item[-] {\scshape Partial-1} \cite{yosinski2014transferable}: Update the last block in the backbone and parameters outside the backbone.

\end{itemize}

\begin{itemize}
\item[$\bullet$] With extra structure: 

%\noindent (The pre-trained layers in these baselines are fixed, and the adapter intermediate dimensions are all 64, following the AdaptFormer (citation)):
%
%\noindent - {\scshape Adapter \cite{houlsby2019parameter}}: Add standard adapter layers after the MSA/MLP layers of each SwinBlock.
%
%\noindent - {\scshape LoRA \cite{hu2021lora}}: Add parallel learnable matrices to multi-head attention weights.
%
%\noindent - {\scshape AdaptFormer \cite{chen2022adaptformer}}: Add parallel adapter layers with scale weights to each MLP layer.
%
%\noindent - {\scshape LoRand \cite{yin20231}}: Add LoRand layers after the MSA/MLP layers of each SwinBlock.

\noindent (The pre-trained layers in these baselines are fixed, and the adapter intermediate dimensions are all 64, following the AdaptFormer \cite{chen2022adaptformer}):

\item[-] {\scshape Adapter} \cite{houlsby2019parameter}: Add standard adapter layers after the MSA/MLP layers of each SwinBlock.

\item[-] {\scshape LoRA} \cite{hu2021lora}: Add parallel learnable matrices to multi-head attention weights.

\item[-] {\scshape AdaptFormer} \cite{chen2022adaptformer}: Add parallel adapter layers with scale weights to each MLP layer.

\item[-] {\scshape LoRand} \cite{yin20231}: Add LoRand++ ($\alpha$=4, $\beta$=16) layers after the MSA/MLP of each SwinBlock. LoRand++ has the best performance among its variants, so the most challenging setting is chosen for comparison.

\end{itemize}

\begin{table*}[tb]
  \caption{\textbf{Results of baselines and our methods on COCO benchmarks.} Swin-B is employed as the pre-trained model here. We present the numbers and percentages of trainable backbone parameters on the left and all the performences on the right. $\ast$ denotes the trainable parameters in backbones. The best AP in each column is bolded.}
  \label{tab:coco}
  \centering
  \scalebox{1}{
		\setlength{\tabcolsep}{2mm}{
			\begin{tabular}{@{}l|rrr|c|cccc@{}}
				\toprule
				\multirow{2}{*}{\textbf{\begin{tabular}[c]{@{}l@{}}\quad Swin-B\\ \quad  (89M)\end{tabular}}} & \multicolumn{1}{c}{\multirow{2}{*}{\textbf{\begin{tabular}[c]{@{}c@{}}Trained$\ast$ \\ Params\end{tabular}}}} & \multicolumn{1}{c}{\multirow{2}{*}{\textbf{\%}}} & 
				\multicolumn{1}{c|}{\multirow{2}{*}{\textbf{$\bm{\Delta_{Full}}$}}} &
				\multicolumn{1}{c|}{\multirow{2}{*}{\textbf{\begin{tabular}[c]{@{}c@{}}Extra\\ Structure\end{tabular}}}} & \multicolumn{4}{c}{\textbf{\begin{tabular}[c]{@{}c@{}}COCO\\ (Cascade Mask R-CNN)\end{tabular}}}  \\ \cline{6-9} 
				
				& \multicolumn{1}{c}{} & \multicolumn{1}{c}{} & \multicolumn{1}{c|}{} &  & \multicolumn{1}{c}{$\bm{\mathrm{AP_{Box}}}$} &\textbf{$\bm{\Delta_{Full}}$} & \multicolumn{1}{c}{$\bm{\mathrm{AP_{Mask}}}$} & \textbf{$\bm{\Delta_{Full}}$} \\ \midrule
				
				\rowcolor{gray!40}\multicolumn{9}{c}{\textit{\textbf{Baselines}}}\\ \midrule
				%------------------------------------------------------------------------------
				\quad {\scshape Full} & \multicolumn{1}{r}{ 89.14 M} & \multicolumn{1}{c}{100.00 \%} &\multicolumn{1}{c|}{-} & \multicolumn{1}{c|}{\ding{55}} & \multicolumn{1}{c}{ 52.40 \%} & \multicolumn{1}{c}{-} & \multicolumn{1}{c}{ 45.10 \%}  & -\\
				\quad {\scshape Fixed} & 0.00 M & 0.00 \% &- 100.00 \% & \ding{55} & \multicolumn{1}{c}{ 48.00 \%} & - 4.40 \%& 41.60 \% & - 3.50 \% \\ 
				\quad {\scshape BitFit} & 0.21 M & 0.23 \% & - 99.77 \%  & \ding{55} & \multicolumn{1}{c}{50.10 \%} & - 2.30 \% & 43.60 \% & - 1.50 \% \\
				\quad {\scshape NormTuning} & 0.06 M & 0.07 \% & - 99.93 \% & \ding{55} & \multicolumn{1}{c}{50.10 \%} & - 2.30 \% & 43.50 \%& - 1.60 \% \\ 
				\quad {\scshape Partial-1} & 12.95 M & 14.53 \% & - 85.47 \% & \ding{55} & \multicolumn{1}{c}{50.60 \%} & - 1.80 \% & 43.70 \%& - 1.40 \% \\ \midrule
				\quad {\scshape Adapter} & 3.19 M & 3.58 \% & - 96.42 \% & \ding{51} & \multicolumn{1}{c}{52.10 \%} & - 0.30 \% & 45.00 \%& - 0.10 \% \\ 
				\quad {\scshape LoRA} & 3.06 M & 3.43 \% & - 96.57 \% & \ding{51} & \multicolumn{1}{c}{50.40 \%} & - 2.00 \% & 43.90 \%& - 1.20 \% \\ 
				\quad {\scshape AdaptFormer} & 1.60 M & 1.79 \% & - 98.21 \% & \ding{51} & \multicolumn{1}{c}{51.70 \%} & - 0.70 \% & 44.60 \%& - 0.50 \% \\ 
				\quad {\scshape LoRand} & 4.68 M & 5.23 \% & - 94.77 \% & \ding{51} & \multicolumn{1}{c}{51.90 \%} & - 0.50 \% & 44.70 \%& - 0.40 \% \\ \midrule

				\rowcolor{gray!40}\multicolumn{9}{c}{\textit{\textbf{Our Method}}}\\ \midrule
				
				\quad {\scshape \textbf{Mona}} & 4.16 M & 4.67 \% & - 95.33 \% & \ding{51} & \textbf{53.40 \%} & \textbf{+ 1.00 \%}& \textbf{46.00} \% & \textbf{+ 0.90 \%} \\ \bottomrule
				
		\end{tabular}
  }}

  % \vspace{-10pt}
\end{table*}

\begin{table*}[tb]
        \caption{\textbf{Results of baselines and our methods on Pascal VOC and ADE20K benchmarks.} Swin-L is employed as the pre-trained model here. We present the numbers and percentages of trainable backbone parameters on the left and all the performences on the right. $\ast$ denotes the trainable parameters in backbones. The best AP/mIoU in each column is bolded.}
	\label{tab:ade}
	\centering
	\scalebox{1}{
		\setlength{\tabcolsep}{2mm}{
			\begin{tabular}{@{}l|rrr|c|cc|cc@{}}
				\toprule
				\multirow{2}{*}{\textbf{\begin{tabular}[c]{@{}l@{}}\quad Swin-L\\ \quad (198M)\end{tabular}}} & \multicolumn{1}{c}{\multirow{2}{*}{\textbf{\begin{tabular}[c]{@{}c@{}}Trained$\ast$ \\ Params \end{tabular}}}} & \multicolumn{1}{c}{\multirow{2}{*}{\textbf{\%}}} & 
				\multicolumn{1}{c|}{\multirow{2}{*}{\textbf{$\bm{\Delta_{Full}}$}}} &
				\multicolumn{1}{c|}{\multirow{2}{*}{\textbf{\begin{tabular}[c]{@{}c@{}}Extra\\ Structure\end{tabular}}}} & \multicolumn{2}{c|}{\textbf{\begin{tabular}[c]{@{}c@{}}Pascal VOC\\ (RetinaNet)\end{tabular}}} & \multicolumn{2}{c}{\textbf{\begin{tabular}[c]{@{}c@{}}ADE20K\\ (UperNet)\end{tabular}}} \\ \cline{6-9} 
				
				& \multicolumn{1}{c}{} & \multicolumn{1}{c}{} & \multicolumn{1}{c|}{} &  & \multicolumn{1}{c}{$\bm{\mathrm{AP_{Box}}}$} &\textbf{$\bm{\Delta_{Full}}$} & \multicolumn{1}{c}{$\bm{\mathrm{mIoU}}$} & \textbf{$\bm{\Delta_{Full}}$} \\ \midrule
				
				\rowcolor{gray!40}\multicolumn{9}{c}{\textit{\textbf{Baselines}}}\\ \midrule
				
				%------------------------------------------------------------------------------
				\quad {\scshape Full} & \multicolumn{1}{r}{ 198.58 M} & \multicolumn{1}{c}{100.00 \%} &\multicolumn{1}{c|}{-} & \multicolumn{1}{c|}{\ding{55}} & \multicolumn{1}{c}{83.70 \%} & \multicolumn{1}{c|}{-} & \multicolumn{1}{c}{ 51.18 \%}  & -\\
				\quad {\scshape Fixed} & 0.00 M & 0.00 \% &- 100.00 \% & \ding{55} & \multicolumn{1}{c}{ 83.80 \%} & + 0.10 \%& 46.84 \% & - 4.34 \% \\ 
				\quad {\scshape BitFit} & 0.30 M & 0.15 \% & - 99.85 \%  & \ding{55} & \multicolumn{1}{c}{85.40 \%} & + 1.70 \% & 48.37 \% & - 2.81 \% \\
				\quad {\scshape NormTuning} & 0.10 M & 0.05 \% & - 99.95 \% & \ding{55} & \multicolumn{1}{c}{85.50 \%} & + 1.80 \% & 47.89 \%& - 3.29 \% \\ 
				\quad {\scshape Partial-1} & 28.77 M & 14.53 \% & - 85.47 \% & \ding{55} & \multicolumn{1}{c}{85.50 \%} & + 1.80 \% & 47.44 \%& - 3.74 \% \\ \midrule
				\quad {\scshape Adapter} & 4.61 M & 2.33 \% & - 97.67 \% & \ding{51} & \multicolumn{1}{c}{86.70 \%} & + 3.00 \% & 50.78 \%& - 0.40 \% \\ 
				\quad {\scshape LoRA} & 4.57 M & 2.31 \% & - 97.69 \% & \ding{51} & \multicolumn{1}{c}{85.40 \%} & + 1.70 \% & 50.34 \%& - 0.84 \% \\ 
				\quad {\scshape AdaptFormer} & 2.34 M & 1.18 \% & - 98.82 \% & \ding{51} & \multicolumn{1}{c}{86.60 \%} & + 2.90 \% & 50.83 \%& - 0.35 \% \\ 
				\quad {\scshape LoRand} & 5.20 M & 2.62 \% & - 97.38 \% & \ding{51} & \multicolumn{1}{c}{86.90 \%} & + 3.20 \% & 50.93 \%& - 0.25 \% \\ \midrule

				\rowcolor{gray!40}\multicolumn{9}{c}{\textit{\textbf{Our Method}}}\\ \midrule
				
				\quad {\scshape \textbf{Mona}} & 5.08 M & 2.56 \% & - 97.44 \% & \ding{51} & \textbf{87.30 \%} & \textbf{+ 3.60 \%}& \textbf{51.36 \%} & \textbf{+ 0.18 \%} \\ \bottomrule
	\end{tabular}}}

		\vspace{-10pt}
\end{table*}

\begin{table}[tb]
  \caption{\textbf{Results on DOTA and STAR benchmarks.} Swin-B is employed as the pre-trained model here. The best AP in each column is bolded.}
  \label{tab:obb}
  \centering
  \scalebox{.86}{
		\setlength{\tabcolsep}{0.5mm}{
			\begin{tabular}{@{}l|cc|c|c@{}}
				\toprule
				\multirow{2}{*}{\textbf{\begin{tabular}[c]{@{}l@{}}\quad Swin-B\\ \quad  (89M)\end{tabular}}} & \multicolumn{2}{c|}{\textbf{\begin{tabular}[c]{@{}c@{}}Oriented R-CNN\\(Faster R-CNN)\end{tabular}}} & \multicolumn{1}{c|}{\textbf{\begin{tabular}[c]{@{}c@{}}KLD\\(RetinaNet)\end{tabular}}} & \multicolumn{1}{c}{\textbf{\begin{tabular}[c]{@{}c@{}}H2RBox-v2 \\(FCOS)\end{tabular}}}   \\ \cline{2-5} 
				
				& \multicolumn{1}{c}{\textbf{DOTA-v1.0}} & \textbf{STAR} &   \textbf{STAR} &  \textbf{STAR}  \\ \midrule
				
				\rowcolor{gray!40}\multicolumn{5}{c}{\textit{\textbf{Baselines}}}\\ \midrule
				%------------------------------------------------------------------------------

                \quad {\scshape Full}                                                                             & 78.31  \%                                   & 38.63 \%                               & 30.33        \%                                                      & 30.29   \%                                                            \\
\quad {\scshape Fixed}                                                                            & 74.10            \%                         & 30.83             \%                   & 23.81    \%                                                          & 26.01       \%                                                        \\
\quad {\scshape BitFit}                                                                           & 76.05      \%                               & 34.51           \%                     & 28.17         \%                                                     & 29.41  \%                                                             \\
\quad {\scshape NormTuning}                                                                      & 75.82                 \%                    & 33.13         \%                        & 27.12            \%                                                   & 27.79            \%                                                    \\
\quad {\scshape Partial-1}                                                                       & 75.72            \%                          & 33.96           \%                      & 28.53                 \%                                              & 28.89    \%                                                            \\ \midrule
\quad {\scshape Adapter}                                                                         & 78.27         \%                             & 37.97                  \%               & 30.35                    \%                                           & 30.24     \%                                                           \\
\quad {\scshape LoRA}                                                                             & 75.91     \%                                 & 33.80               \%                  & 27.48         \%                                                      & 28.95     \%                                                           \\
\quad {\scshape AdaptFormer}                                                                      & 77.43     \%                                 & 35.95              \%                   & 29.36   \%                                                            & 30.11         \%                                                       \\
\quad {\scshape LoRand}                                                                           & 77.65        \%                              & 36.44           \%                      & 29.83       \%                                                        & 28.85  \%                                                              \\ \midrule
				% \quad {\scshape Full} & 72.57 & 30.33 & 74.70 & 30.29 & & 31.16 & 78.31 & 38.63 \\
				% \quad {\scshape Fixed} & 64.51 & 23.81 & 67.55 & 26.01 & & 27.91 & 74.10 & 30.83 \\ 
				% \quad {\scshape BitFit} & 68.73 & 28.17 & 69.95 & 29.41 & & 28.25 & 76.05 & 34.51 \\
				% \quad {\scshape NormTuning} & 69.47 & 27.12 & 69.44 & 27.79 & & 26.79 & 75.82 & 33.13  \\ 
				% \quad {\scshape Partial-1} & 67.18 & 28.53 & 69.64 & 28.89 & & 26.39 & 75.72 & 33.96  \\ \midrule
				% \quad {\scshape Adapter} & 69.39 & 30.35 & 72.79 & 30.24 & & 28.93 & 78.27 & 37.97  \\ 
				% \quad {\scshape LoRA} & 64.78 & 27.48 & 70.15 & 28.95 & & 26.85 & 75.91 & 33.80 \\ 
				% \quad {\scshape AdaptFormer} & 67.78 & 29.36 & 71.61 & 30.11 & & 28.19 & 77.43 & 35.95 \\ 
				% \quad {\scshape LoRand} & 69.55 & 29.83 & 69.95 & 28.85 & & 28.42 & 77.65 & 36.44 \\ \midrule
				
				\rowcolor{gray!40}\multicolumn{5}{c}{\textit{\textbf{Our Method}}}\\ \midrule
				
				\quad {\scshape \textbf{Mona} }                                                                   & \textbf{78.44 \% }                            & \textbf{39.45 \% }                       & \textbf{30.90 \% }                                                     & \textbf{31.34 \% }                                  \\ \bottomrule
				
		\end{tabular}
  }}

  % \vspace{-20pt}
\end{table}

\begin{table*}[tb]
        \caption{\textbf{Results of baselines and our methods on three classification datasets.} Swin-L is employed as the pre-trained model here. We present top-1 accuracy (\%) and top-5 accuracy (\%) of each dataset. The best result in each column is bolded.}
	\label{tab:cls}
	\centering
	\scalebox{1}{
		\setlength{\tabcolsep}{2mm}{
			\begin{tabular}{@{}lcccccccc@{}}
				\toprule
				\multirow{2}{*}{\quad \textbf{Method}} & \multicolumn{2}{c}{\textbf{Flowers102}}  & \multicolumn{2}{c}{\textbf{OxfordPets}}  & \multicolumn{2}{c}{\textbf{VOC2007} } & \multicolumn{2}{c}{\textbf{Average}} \\ \cmidrule(l){2-9} 
				& top-1 acc. & top-5 acc. & top-1 acc. & top-5 acc. & top-1 acc.     & top-5 acc. & top-1 acc.     & top-5 acc.    \\ \midrule
				
				\rowcolor{gray!30}\multicolumn{9}{c}{\textit{\textbf{Baselines}}}\\ \midrule
				
				\quad {\scshape Full}         &  99.5772  &  99.8536  &  94.6579  &  99.6257   &  84.1276  &  96.9507  & 92.7876 & 98.8100 \\
				\quad {\scshape Fixed}        &  99.3007  &  99.8374  &  94.2219  &  99.9182   &  85.0162  &  98.9499  & 92.8463 & 99.5685 \\
				\quad {\scshape BitFit}       &  99.5772  &  99.8211  &  95.3393  &  99.9182   &  85.6018  &  99.3336  & 93.5061 & 99.6910 \\
				\quad {\scshape NormTuning}   &  99.5284  &  99.8374  &  95.2303  &  99.8910   &  85.5210  &  99.2528  & 93.4266 & 99.6604 \\
				\quad {\scshape Partial-1}    &  99.6585  &  99.8374  &  95.3938  &  99.8637   &  84.9354  &  98.6066  & 93.3292 & 99.4359 \\
				\quad {\scshape Adapter}      &  99.5934  &  99.8536  &  95.3393  &  99.8092   &  \textbf{87.0355}  &  99.1317  & 93.9894 & 99.6144 \\
				\quad {\scshape LoRA}         &  99.5446  &  99.8536  &  95.1485  &  99.8910   &  85.7028  &  99.3134  & 93.4653 & 99.6860 \\
				\quad {\scshape AdaptFormer}  &  99.5609  &  99.8536  &  95.2576  &  99.8365   &  86.2884  &  99.2730  & 93.7023 & 99.6544 \\
				\quad {\scshape LoRand}       &  99.5725  &  99.8536  &  95.3515  &  99.8910   &  86.6534  &  99.3741  & 93.8591 & 99.7062 \\ \midrule
				\rowcolor{gray!30}\multicolumn{9}{c}{\textit{\textbf{Our Method}}}\\ \midrule
				\quad {\scshape \textbf{Mona}}&  \textbf{99.6764}  &  \textbf{99.9024}  &  \textbf{95.4765}  & \textbf{99.9182}  &  86.9709
				&  \textbf{99.5057} & \textbf{94.0413}&\textbf{99.7592 
				}\\ \bottomrule
				
	\end{tabular}}}
	\vspace{-10pt}
\end{table*}

\subsection{Main Results}
\label{results}

Instance segmentation on COCO is challenging. From Table \ref{tab:coco}, we find that Mona outperforms all PEFT baselines and is the only method that outperforms full fine-tuning even by 1\%. COCO experiments effectively demonstrate the capability of the proposed method and show a better option than full fine-tuning in terms of storage and performance. Among delta-tuning methods, most baselines without extra structure can save more new parameters (except Partial-1), but their average performance is lower than that with extra structure. For baselines with additional structure, adapter-based approaches is superior to the reparameterization-based LoRA. Table \ref{tab:coco} shows that LoRA performs well on NLP tasks but poorly on computer vision tasks. Table \ref{tab:coco} indicates that the performance of delta-tuning is not directly related to parameter sizes. Partial-1 has the most trainable parameters, but its performance is significantly lower than that of adapter-based baselines. This result suggests that superior module design can effectively enhance the transferring efficiency of pre-trained models and reduce massive new parameters.

Table \ref{tab:ade} shows the results of Pascal VOC (object detection) and ADE20K (semantic segmentation). Also, Mona outperforms all other methods in Table \ref{tab:ade}. Mona outperforms full fine-tuning by 3.6\% and 0.18\% on these two tasks. Table \ref{tab:ade} again indicates that full fine-tuning is not the best choice for visual transfer learning. Interestingly, all baselines surpasses full fine-tuning on VOC, which is different from COCO and ADE20K. Relatively little data in VOC may lead to over-fitting when full fine-tuning a 198M Swin-Large pre-trained model. Compared to full fine-tuning, other methods fix most pre-trained parameters, so the model performance is less likely to collapse severely during tuning. NLP scholars treat similar cases as low-resource cases \cite{dodge2020fine, peters2019tune}. VOC here can be considered as a low-resource case in CV. For ADE20K, the performance gaps between baselines without additional structure and adapter-based baselines are more significant than VOC and COCO. For parameter sizes, most methods in Tables \ref{tab:coco} and \ref{tab:ade} (except Partial-1) produce less than 5\% new backbone parameters, which is the characteristic of delta-tuning. Despite the slight increase in parameters, Mona still outperforms the previous art and breaks the full fine-tuning performance ceiling by a wide margin.

Table \ref{tab:obb} shows the performance on the more challenging oriented object detection tasks. Firstly, Columns 2-3 of Table \ref{tab:obb} show that Mona outperforms full fine-tuning and other efficient fine-tuning methods on two datasets with Oriented R-CNN \cite{xie2021oriented}. Secondly, STAR has more instances and classes than DOTA, which is more challenging than DOTA. So we experiment with STAR on more frameworks. Columns 4-5 show the results of all methods with KLD \cite{yang2021learning} and H2RBox-v2 \cite{yu2024h2rbox}. It can be seen that Mona outperforms all baseline methods on these frameworks. Table \ref{tab:obb} further illustrates that the proposed adapter can lead to performance breakthroughs on a wider range of visual tasks.

For classification tasks, we show the individual and average results on three classification datasets in Table \ref{tab:cls}. Mona outperforms all the baselines on Flowers102, OxfordPets, and outperforms the average results of all baselines. Table \ref{tab:cls} indicates that Mona has a high transfer efficiency on relatively simple tasks. In addition, we find that the average results of all delta-tuning methods surpass full fine-tuning, which is similar to conclusions in previous art \cite{he2022parameter}. Compared to classification, the complex dense prediction tasks can demonstrate the advantages and disadvantages of different tuning approaches more intuitively.

%-----------------12点39分--------------------

In summary, the results of Tables \ref{tab:coco} to \ref{tab:obb} can be summarized in two aspects: 1) As to performance, the widely used full fine-tuning paradigm in art like Swin is no longer the optimal choice for visual tasks. The proposed Mona-tuning surpasses the performance ceiling of full fine-tuning in representative tasks such as instance segmentation, semantic segmentation, object detection, image classification, and oriented object detection. Specifically, Mona achieves a 1\% AP gain over full fine-tuning in the challenging COCO instance segmentation task. 2) Mona, based on multi-cognitive visual filtering, surpasses recent remarkable baselines in most tasks. Mona comprehensively enhances the practicality and generality of delta-tuning in visual tasks. Mona-tuning not only significantly reduces storage costs, but also further elevates the performance ceiling of visual tasks.

\begin{figure}[tb]
	\centering
	\includegraphics[scale=.30]{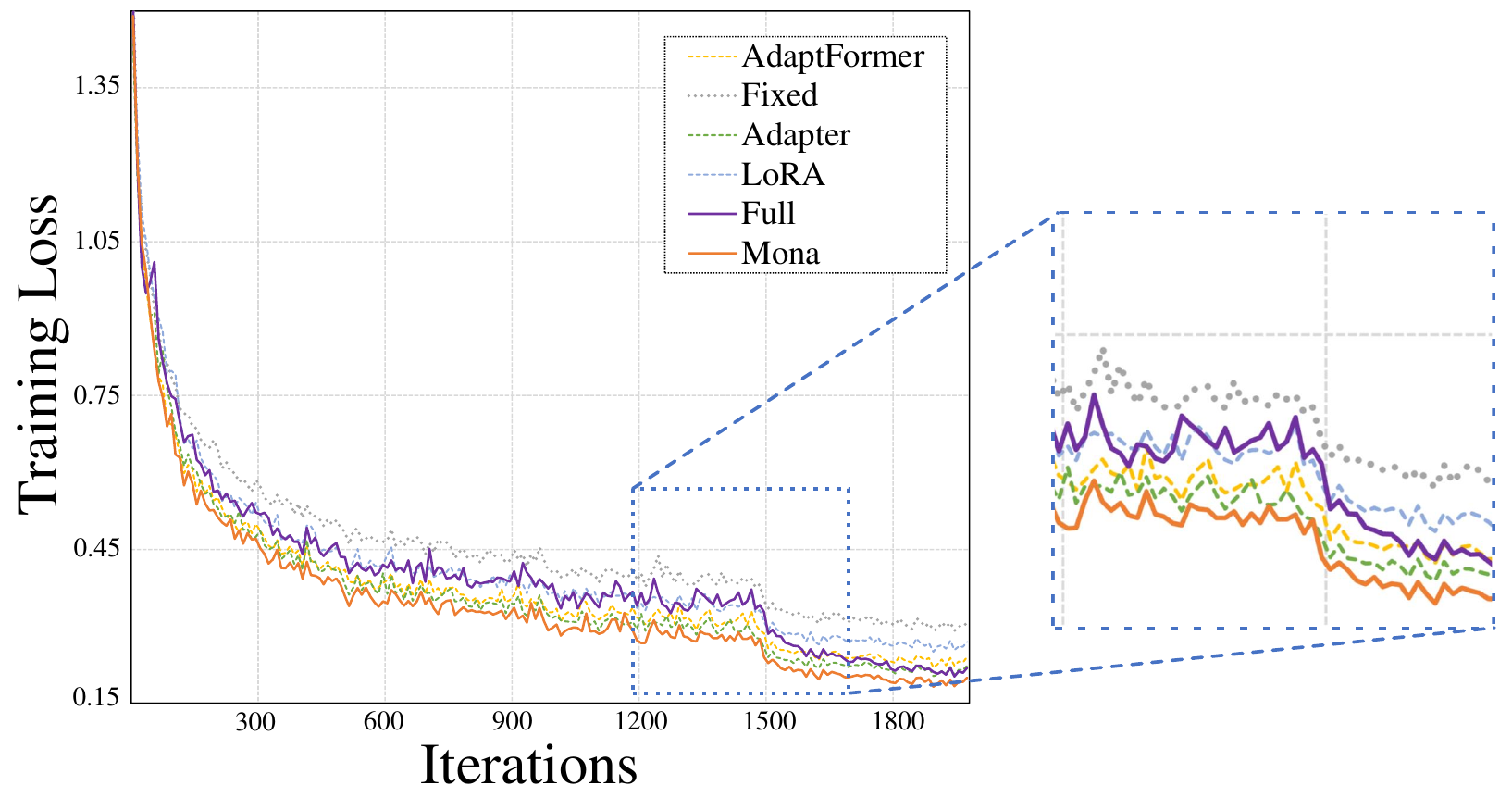}
	\caption{\textbf{Loss curves.} Among all the methods, the proposed method converges faster and significantly exceeds the full fine-tuning.} \label{fig:loss}

\vspace{-10pt}
\end{figure}

\subsection{Loss Analysis}
\label{loss}

We present the loss converging process for Mona and five representative  baselines on the object detection task (Pascal VOC) in Figure \ref{fig:loss}. The proposed method yields a significant advantage in the convergence process compared to full fine-tuning, which explains its better performance on VOC. Mona also converges faster than other delta-tuning methods, suggesting that multi-cognitive visual filters can better process visual features and accelerate the convergence of transfer learning. Convergence analysis again demonstrates that the proposed method is a highly competitive visual transfer learning method and full fine-tuning is no longer the optimal choice for visual tasks.

\subsection{Ablations}
\label{ablations}

In this section, we ablate multiple potential factors that affect model performance, including 
% the impact of ``Average'' and ``Scaled LayerNorm'' designs, 
intermediate dimensions of adapters and model sizes
% , and frameworks. 
All ablation experiments are conducted on Pascal VOC.

% \begin{table}[h]
%         \caption{\textbf{Ablations of two designs.} The results show that both designs provide positive improvements to the structure, and better results are obtained when they are present together.}
%         \label{tab:ab-design}
% 	\centering
% 	\scalebox{1}{\setlength{\tabcolsep}{2.5mm}{
% 	\begin{tabular}{@{}ccc@{}}
% 		\toprule
% 		\textbf{Scaled LayerNorm} & \textbf{Average} &$\bm{\mathrm{AP_{Box}}}$ \\ \midrule
% 		\ding{55}        & \ding{55}                & 86.7\%      \\
% 		\ding{51}        & \ding{55}                & 87.1\%      \\
% 		\ding{55}        & \ding{51}                & 86.8\%      \\
% 		\ding{51}        & \ding{51}                & 87.3\%      \\ \bottomrule
% 	\end{tabular}}}
%  % \vspace{-5pt}
% \end{table}

% Two detail designs are included in the iterative design section, namely scaled LayerNorm and Averaging operation. We show their ablation experiments in Table \ref{tab:ab-design}. The results show that both designs improve the performance of Mona-tuning, and scaled LayerNorm achieves higher gains. Moreover, Mona-tuning achieves better performance when both designs are present at the same time. Table \ref{tab:ab-design} demonstrates the necessity and effectiveness of design iterations.

\begin{table}[tb]
        \caption{\textbf{Ablations of intermediate dimensions.} 64 intermediate dimensions achieves the best performance. $\ast$ denotes the trainable parameters in backbones.}
        \label{tab:ab-dim}
	\centering
	\scalebox{1}{\setlength{\tabcolsep}{2.5mm}{
	\begin{tabular}{@{}ccc@{}}
		\toprule
		\begin{tabular}[c]{@{}c@{}} \textbf{Intermediate} \\ \textbf{Dimensions}\end{tabular} & \multicolumn{1}{l}{\begin{tabular}[c]{@{}l@{}}\textbf{Trained}\\ \textbf{Params*}\end{tabular}} & $\bm{\mathrm{AP_{Box}}}$   \\ \midrule
		32   & 1.35 \%      & 86.8 \%\\
		64   & 2.56 \%      & 87.3 \%\\
		128  & 5.22 \%      & 87.1 \%\\ \bottomrule
	\end{tabular}}}
\vspace{-15pt}
\end{table}

The workflow of the adapter is to compress the input from pre-trained layers into a low-dimensional feature space and transfer the pre-trained knowledge by tuning adapters. Thus, the intermediate dimension is important for adapters. We ablate the intermediate dimension of Mona in Table \ref{tab:ab-dim} and fix other settings. Dimension candidates are 32, 64, and 128. Table \ref{tab:ab-dim} show that the 64-dimension surpasses 32-dimension and 128-dimension. Chen et al. \cite{chen2022adaptformer} also study the intermediate dimension of AdaptFormer. They find that the 64-dimension AdaptFormer surpasses its 32- and 256-dimension versions in visual classification tasks, which is consistent with our conclusion. The results of Table \ref{tab:ab-dim} and Chen et al. indicate that the intermediate dimension of the adapter is not proportional to the performance, which means that a larger number of adapter parameters does not necessarily lead to better results. 

% \begin{table}[h]
%         \caption{\textbf{Results on PVT.} Mona has impressive versatility, which also shows strengths in other frameworks.}
%         \label{tab:pvt}
% 	\centering
% 	\scalebox{1}{\setlength{\tabcolsep}{2mm}{
% 	\begin{tabular}{@{}lc@{}}
% 		\toprule
% 		\textbf{Methods}      & $\bm{\mathrm{AP_{Box}}}$ \\ \midrule
% 		Full (PVT-L) & 76.1\%      \\
%             Adapter (PVT-L) & 78.9\%      \\
%             LoRA (PVT-L) & 77.6\%      \\
%             AdaptFormer (PVT-L) & 79.2\%      \\
%             LoRand (PVT-L) & 79.3\%      \\
% 		Mona (PVT-L) & 80.3\%      \\ \bottomrule
% 	\end{tabular}}}
% % \vspace{-10pt}
% \end{table}

% After that, we conduct experiments on PVT \cite{wang2021pyramid} to test the generality of Mona. Table \ref{tab:pvt} shows that Mona outperforms not only full fine-tuning but also other baselines on PVT (PVT-Large), which indicates that Mona is effective on a wider range of frameworks.

% \usepackage{booktabs}
\begin{table}[h]
        \caption{\textbf{Performance of mona on models with different sizes.} The results indicate that model sizes do not constrain Mona's superiority.}
	\label{tab:ab-model}
	\centering
	\scalebox{1}{\setlength{\tabcolsep}{2mm}{
	\begin{tabular}{@{}lccc@{}}
		\toprule
		\textbf{Model} &\begin{tabular}[c]{@{}c@{}}{\scshape \textbf{Full}}\\ \textbf{(VOC)}\end{tabular} & \begin{tabular}[c]{@{}c@{}}{\scshape \textbf{Mona}}\\ \textbf{(VOC)}\end{tabular} & \multicolumn{1}{c}{\begin{tabular}[c]{@{}c@{}}\textbf{Param \%}\\ \textbf{(Mona)}\end{tabular}} \\ \midrule
		Swin-T & 80.1 \%& 83.5 \% & 4.87 \%                                                                        \\
		Swin-B & 81.6 \% & 86.5 \%& 4.06 \%                                                                        \\
		Swin-L & 83.7 \%& 87.3 \%& 2.56 \%                                                                        \\ \bottomrule
	\end{tabular}}}
	% \vspace{-10pt}
\end{table}

In Table \ref{tab:ab-model}, we change the size of the backbone networks under the same settings, and the model candidates are 29M Swin-T, 88M Swin-B, and 197M Swin-L. We can draw the following three conclusions from Table \ref{tab:ab-model}. First, the more parameters the backbone network has, the smaller the proportion of Mona parameters for the same Mona setting. This result indicates that Mona-tuning can save more parameters when the backbone gets larger. Existing visual models are getting larger and larger. InternImage-H \cite{wang2023internimage} reaches 1.08B parameters, and SwinV2-G \cite{liu2022swin} reaches 3B. Parameter-efficient Mona-tuning can save billions of parameters and massive storage costs in the era of large models. Second, Mona surpasses full fine-tuning on three model settings, and its performance improves when model size grows. Table \ref{tab:ab-model} shows that Mona-tuning can improve training efficiency and performance in smaller models. We just discussed Mona's advantages for large models. However, more resource-limited research teams and project groups use small models. Mona-tuning also has the potential to help resource-limited researchers leverage high-performance large models in their own applications. Third, the proposed method is more capable of stimulating the potential of large models compared to full fine-tuning. From Swin-T to Swin-L, full fine-tuning brings 3.6\% performance gain, while Mona brings 3.8\%. In other words, Mona can perform better as the model gets larger and help further increase the upper bound for performance-sensitive tasks.

% \begin{minipage}[c]{0\textwidth}
% \centering
% \scalebox{.95}{\setlength{\tabcolsep}{.1mm}{\begin{tabular}{@{}ccc@{}}
% 		\toprule
  
% 		\begin{tabular}[c]{@{}c@{}} \textbf{Inter.} \\ \textbf{Dim.}\end{tabular} & \multicolumn{1}{l}{\begin{tabular}[c]{@{}l@{}}\textbf{Trained}\\ \textbf{Params*}\end{tabular}} & $\bm{\mathrm{AP_{Box}}}$   \\ \midrule
% 		32   & 1.35 \%      & 86.8 \%\\
% 		64   & 2.56 \%      & 87.3 \%\\
% 		128  & 5.22 \%      & 87.1 \%\\ \bottomrule
% 	\end{tabular}}}
% \captionof{table}{\textbf{Ablations of inter. dim.}}
% \end{minipage}
% \begin{minipage}[c]{.7\textwidth}
% \centering
% \scalebox{.75}{\setlength{\tabcolsep}{2mm}{\begin{tabular}{@{}lc@{}}
% 		\toprule
% 		\textbf{Methods}      & $\bm{\mathrm{AP_{Box}}}$ \\ \midrule
% 		Full (PVT-L) & 76.1\%      \\
%             Adapter (PVT-L) & 78.9\%      \\
%             LoRA (PVT-L) & 77.6\%      \\
%             AdaptFormer (PVT-L) & 79.2\%      \\
%             LoRand (PVT-L) & 79.3\%      \\
% 		Mona (PVT-L) & 80.3\%      \\ \bottomrule
% 	\end{tabular}}}
% \captionof{table}{\textbf{Results on PVT.}}
% \end{minipage}

% \usepackage{booktabs}

\section{Discussion}
Inference cost is a concern for the industry. Methods based on adapter modules (such as Adapter, AdaptFormer, LoRand, and Mona) introduce additional structures, along with a small increase in inference cost. Re-parameterization-based approaches (like LoRA) do not affect inference cost but tend to perform poorly on visual tasks. The utility of adapters could be greatly enhanced by combining the strengths of both approaches. We will continue to strive towards achieving this goal.

\section{Conclusion}
This paper propose a novel visual fine-tuning method, the multi-cognitive visual adapter (Mona) tuning, which effectively enhances the efficiency and performance of visual fine-tuning. Comprehensive experiments demonstrate that the proposed Mona outperforms traditional full fine-tuning paradigms and other delta-tuning methods across representative tasks, including instance segmentation, semantic segmentation, object detection, image classification, and oriented object detection. In the era of large models, full fine-tuning is no longer the optimal choice for visual tasks. We hope that Mona-tuning can improve the knowledge transferring efficiency of large models and bring performance breakthroughs on more visual tasks.

\bibliography{aaai25}

\end{document}